\documentclass[12pt]{article}

\usepackage{latexsym}

\makeatletter
\def\newleaf{\newpage
\newcount\tmp
\tmp=\c@page
\divide\tmp by 2
\multiply\tmp by 2
\ifnum\c@page=\tmp
~\newpage
\fi
}
\makeatother

\expandafter\ifx\csname proofnewpage\endcsname\relax

\fi

\def\color[#1]#2{}

\long\def\nop#1{}

\def\comment{\edef\cps{\the\parskip} \parskip=0.5cm \begingroup \tt}

\expandafter\ifx\csname shortcite\endcsname\relax

\fi

\newbox\current

\long\def\plframebox#1{
\setbox\current\vbox{#1}		

\vbox to \ht\current {\hrule\vss
\hbox to \wd\current {%
\vrule \hss\box\current\hss \vrule}
\vss\hrule }
}

\long\def\eatpar#1{%
\ifx#1\par                      
\let\nextmove=\eatpar           
\else
\let\nextmove=#1
\fi
\noexpand\nextmove
}

\def\modifymargins#1#2{
\newdimen\addtoh
\newdimen\addtow
\addtoh=#1
\addtow=#2

\advance\topmargin by -\addtoh
\multiply\addtoh by 2
\advance\textheight by \addtoh

\advance\oddsidemargin by -\addtow
\advance\evensidemargin by -\addtow
\multiply\addtow by 2
\advance\textwidth by \addtow
}

\begingroup
\catcode`\~=11
\gdef\centertilde#1{\lower #1pt\hbox{~}}
\endgroup

\newcount\currenttime
\newcount\hour
\newcount\minute

\def\printtime{%
\currenttime=\time
\hour=\currenttime
\divide\hour by 60
\minute=-\hour
\multiply\minute by 60
\advance\minute by \currenttime
\the\hour:\ifnum\minute<10 0\fi\the\minute
}

\begingroup
\makeatletter
\global\let\@@date=\@date
\gdef\@date{\@@date\ --- \printtime}
\endgroup

\def\oggi{\number\day\space 
\ifcase\month\or
Gennaio\or Febbraio\or Marzo\or Aprile\or Maggio\or Giugno\or
Luglio\or Agosto\or Settembre\or Ottobre\or Novembre\or Dicembre\fi
\space \number\year}

\newcounter{rmexample}

\def\proof{\noindent {\sl Proof.\ \ }}

\def\qed{\hfill{\boxit{}}
  \ifdim\lastskip<\medskipamount \removelastskip\penalty55\medskip\fi}
\def\qedn#1{\hfill{\boxit{}$_#1$}
  \ifdim\lastskip<\medskipamount \removelastskip\penalty55\medskip\fi}
\long\def\boxit#1{\vbox{\hrule\hbox{\vrule\kern3pt
                  \vbox{\kern3pt#1\kern3pt}\kern3pt\vrule}\hrule}}

  \def\D{{\cal D}}

\def\l{\langle}
\def\r{\rangle}

\def\true{{\sf true}}
\def\false{{\sf false}}

\def\np{{\rm NP}}
\def\conp{{\rm coNP}}

\def\S#1{\mbox{$\Sigma^p_{#1}$}}
\def\P#1{\mbox{$\Pi^p_{#1}$}}
\def\D#1{\mbox{$\Delta^p_{#1}$}}
\def\Dlog#1{\mbox{$\Delta^p_{#1}[\log n]$}}

\def\pspace{{\rm PSPACE}}

\def\profont{\sf}

\def\x3c{{\profont x3c}}

\def\possnewtheorem#1#2{
\expandafter\ifx\csname #1\endcsname\relax
\newtheorem{#1}{#2}
\fi
}

\def\possnewtheoremthree#1[#2]#3{
\expandafter\ifx\csname #1\endcsname\relax
\newtheorem{#1}[#2]{#3}
\fi
}

\possnewtheorem{theorem}{Theorem}
\possnewtheorem{corollary}{Corollary}
\possnewtheorem{lemma}{Lemma}
\possnewtheoremthree{proposition}[theorem]{Proposition}
\possnewtheorem{definition}{Definition}
\possnewtheorem{question}{Question}
\possnewtheorem{example}{Example}
\possnewtheorem{nontheorem}{Counterexample}
\possnewtheorem{property}{Property}
\possnewtheorem{assumption}{Assumption}
\possnewtheorem{conjecture}{Conjecture}
\possnewtheorem{notation}{Notation}
\newenvironment{theorem*}[1]{{\noindent \bf Theorem~#1}\begin{it}}{\end{it}\

}

\modifymargins{0pt}{30pt}

\title{Raising a Hardness Result}
\author{Paolo Liberatore%
\thanks{Dipartimento di Informatica e Sistemistica Antonio Ruberti,
Universit\`a di Roma ``La Sapienza'',
Via Ariosto 25, 00185, Roma, Italy.
Email: paolo@liberatore.org}
}

\begin{document}

\maketitle

\begin{abstract}

This article presents a technique for proving problems hard
for classes of the polynomial hierarchy or for \pspace. The
rationale of this technique is that some problem
restrictions are able to simulate existential or universal
quantifiers. If this is the case, reductions from Quantified
Boolean Formulae (QBF) to these restrictions can be
transformed into reductions from QBFs having one more
quantifier in the front. This means that a proof of hardness
of a problem at level n in the polynomial hierarchy can be
split into n separate proofs, which may be simpler than a
proof directly showing a reduction from a class of QBFs to
the considered problem.

\end{abstract}
 %

\let\sectionnewpage=\relax
\tableofcontents
\let\sectionnewpage=\newpage

\section{Introduction}

Several logics-related problems are complete for classes of
the polynomial hierarchy other than \np\  and \conp. This is
because these problem involve logical consistency and
entailment, which are already \np-hard and \conp-hard,
respectively. Therefore, problems that are defined in terms
of a number of consistency or entailment verifications may
be complete for classes such as \Dlog{2}, \D{2}, or even for
higher classes when this number is exponential. An example
of a problem that can be formulated in terms of an
exponential number of such operations is that of checking
the existence of a formula that is equivalent to a given one
and of size bounded by a given number $k$. This problem can
indeed be expressed as follows: if $k$ is larger than the
given formula, the answer is "yes"; otherwise, guess a
formula of size bounded by $k$ and check equivalence (mutual
entailment) to the given formula. This problem is therefore
in \S{2}. Hardness to the same class has been proved in
particular cases~\cite{hema-wech-97}.

Several results of \S{2}-hardness and \P{2}-hardness have
been published since the beginning of the 90s
\cite{gott-92-b,stil-92,eite-gott-92-d,eite-gott-93-b,cado-lenz-94,cado-doni-scha-94}.
Some problems have even been proved to be hard for classes
at the fourth level of the polynomial hierarchy
\cite{eite-gott-96,eite-gott-leon-97,dimo-nebe-toni-02,eite-tomp-wolt-05}.

For example, the problem of relevance in skeptical default
abduction based on minimal explanations is \S{4}-complete.
This has been proved by Eiter, Gottlob, and
Leone~\cite{eite-gott-leon-97} by showing a reduction such
that a QBF in the form $\exists X \forall Y \exists Z
\forall K.F$ is valid if and only if a certain hypothesis is
in some minimal explanations of a certain problem of
skeptical default abduction. Such a proof is relatively
complicated, as it requires showing that, if there exists an
evaluation of the variables $X$ such that for all possible
evaluation of the variables $Y$, etc. then the hypothesis is
in some minimal explanation, and vice versa. As a
comparison, a proof of \np-hardness done by reduction from
propositional satisfiability only involves an evaluation of
the variables in~$X$.

The technique proposed in this article simplifies such
proofs by requiring only one quantifier at time to be
considered. This may not necessarily simplify the search for
a reduction, but allows its formal proof to be split into a
number of simpler sub-proofs.

In particular, the idea is to start from the assumption that
a reduction from QBF to a given problem works, and show that
this reduction can be "raised", that is, modify in such a
way the QBF has a single more quantifier in the front. For
example, assume that a problem has been proved \P{3}-hard by
a reduction from QBFs in the form $\forall Y \exists Z
\forall K.F$ to a given problem is already known. In some
cases, we can use this reduction to produce a new reduction
from QBFs in the form $\exists x \forall Y \exists Z \forall
K.E$ to the problem under consideration, where $x$ is a new
variable. If this step can be iterated for a polynomial
number of times, that would result in a proof of
\S{4}-hardness. The (iterated) addition of a single
quantifier raised a \P{3}-hardness proof to a \S{4}-hardness
proof.

Formally, let $P$ be problem under consideration, and assume
that $I$ is a translation from a class of QBFs into $P$.
This means that every formula $Q.E$, where $Q$ is a sequence
of quantifiers and $E$ a propositional formula, is
translated into an instance $I(Q,E)$ of $P$ such that:

\begin{eqnarray*}
Q.E \mbox{ is valid } & \mbox{iff} & I(Q,E) \in P
\end{eqnarray*}

For any formula $F$ containing a variable $x$, we let
$F|_{x=v}$, where $x$ is a variable and $v$ is either
$\true$ or $\false$, be the formula obtained by replacing
each occurrence of $x$ with $v$. By this replacement,
$F|_{x=v}$ does not contain the variable $x$. For example,
$((y \wedge \neg x) \vee x)|_{x=\true}$ is $(y \wedge \neg
\true) \vee \true$. This formula is equivalent to $\true$
but syntactically different to it.

If $F$ is a propositional formula containing only variables
in $Q$ and another variable $x$ which is not in $Q$, then
both $Q.F|_{x=\true}$ and $Q.F|_{x=\false}$ are well-formed
QBF formulae, since both $F|_{x=\true}$ and $F|_{x=\false}$
only contains variables in $Q$. Note that neither
$Q.F|_{x=\true}$ nor $Q.F|_{x=\false}$ contain the variable
$x$. Since $Q.F|_{x=\true}$ and $Q.F|_{x=\false}$ have the
same sequence of quantifiers $Q$, they can both be
translated to $P$:

\begin{eqnarray*}
Q.F|_{x=\true} \mbox{ is valid } & \mbox{iff} & I(Q,F|_{x=\true}) \in P
\\
Q.F|_{x=\false} \mbox{ is valid } & \mbox{iff} & I(Q,F|_{x=\false}) \in P
\end{eqnarray*}

Depending on the reduction $I$, the two instances
$I(Q,F|_{x=\true})$ and $I(Q,F|_{x=\false})$ may be similar.
If this is the case, one can try to merge them into a single
instance $I'(Q,F,x)$ such that:

\begin{eqnarray}
\label{merge-instances}
I'(Q,F,x) \in P \mbox{~~~~iff~~~~}
\left\{
\begin{array}{l}
I(Q,F|_{x=\true}) \in P \\
\hfil\mbox{or} \\
I(Q,F|_{x=\false}) \in P
\end{array}
\right.
\end{eqnarray}

If such a merge is possible, it produces an instance
$I'(Q,F,x)$ which is in $P$ if and only if $\exists x Q.F$
is valid (note that $F$ contains $x$):

\[
\exists x Q.F \mbox{~~~~iff~~~~} I'(Q,F,x) \in P
\]

If this step can be iterated a linear number of times while
not super-polynomially increasing the size of the generated
problem instances, then one has a reduction from the
validity of $\exists X Q.F$ to $P$.

The key to the proof is Equation~\ref{merge-instances}: the
two instances of $P$ can be merged. These two instances
$I(Q,F|_{x=\true})$ and $I(Q,F|_{x=\true})$ are not
arbitrary instances but the result of translating two QBFs
with the same quantifier and similar matrixes.

This is often possible. In practice, many translations from
QBFs to logic-based problems use the matrix of the QBF ``as
is'', by simply copying it verbatim in some part of the
instance of the problem $P$. In this case,
$I(Q,F|_{x=\true})$ and $I(Q,F|_{x=\false})$ are the same
except for the part containing the matrix, where they only
differ because one contains $F|_{x=\true}$ and the other
contains $F|_{x=\false}$. In some cases, these two instances
can be merged by simply taking $I(Q,F|_{x=\true})$,
replacing $F|_{x=\true}$ with $F$, and minimally modifying
the rest of the instance in such a way the answer can be
expressed in terms of the answers to the two subproblems
obtained by setting $x=\true$ and $x=\false$.

This merge needs not only to be possible, but also to
generate instances such that merging can be applied again.
If this iteration is possible while keeping the instance
size polynomial, hardness can be raised of one level in the
polynomial hierarchy. Consider a problem that has been
proved \P{n}-hard by a reduction from QBF to it. By
iterating the step of adding an existential quantifier, one
obtains a proof of \S{n+1}-hardness for the same problem.

This whole process may appear complicated at first, but is
actually easier to perform to specific problems than to
explain in its general form. One thing that one may easy
overlook when considering specific problems is that of
assuming that $Q.F$ is a QBF. This is not the case, as $F$
contains the variable $x$, which is not in $Q$; as a result,
$Q.F$ is not a well-formed QBF. The QBFs mentioned in the
proof are $Q.F|_{x=\true}$, $Q.F|_{x=\false}$, and $\exists
xQ.F$. This is also reflected in the problem instances:
$I(Q,F|_{x=\true})$ and $I(Q,F|_{x=\false})$ do not contain
the variable $x$ because $x$ is not mentioned in $Q$,
$F|_{x=\true}$, and $F_{x=\false}$ (in the latter two
formulae $x$ is replaced by $\true$ and $\false$,
respectively.) The variable $x$ only occurs in the instance
$I'(Q,F,x)$.

A similar method can be used to prove that a universal
quantifier can be added in front of a QBF. The only part
that changes is that the merged instance is in $P$ if both
the two original instances are in $P$. In other words, the
``or'' in the right-hand size of
Equation~\ref{merge-instances} is replaced by ``and''.

In the following sections, we apply this technique to
logic-based abduction
\cite{peir-55,byla-etal-91,eite-gott-95-a,cial-96,eite-gott-leon-97,eite-maki-02,libe-scha-07},
default logic
\cite{reit-80,gott-92-b,cado-scha-93,anto-99,baum-gott-02},
and planning
\cite{fike-nils-71,back-nebe-92,byla-94,back-jons-95,koeh-96}.
In the first case, we add existential quantifiers, in the
second universal quantifiers, and in the third both kinds.

 %

\section{Logic-Based Abduction}

We consider the problem of checking the existence of
explanations in logic-based abduction. This is in essence
the problem of making hypotheses over the possible causes of
observable manifestations \cite{peir-55}. Formally, an
instance of the problem of logic-based abduction is a triple
$\l H,M,T \r$, where $H$ is a set of propositional variables
(hypothesis), $M$ another set of propositional variables
(manifestations), and $T$ a propositional formula relating
$H$ and $M$. An explanation is a subset $S \subseteq H$ such
that $S \cup T$ is consistent and $S \cup T \models M$.
Checking whether an explanation for an instance $\l H,M,T
\r$ exists is \S{2}-complete \cite{eite-gott-95-a}. We
provide an alternative proof using raising.

The starting point is a proof of \conp-hardness, which is
easy to give: a formula $E$ is inconsistent if and only if
the following problem has explanations: hypotheses
$H=\emptyset$, manifestations $M=\{a\}$, theory $T=\{\neg E
\vee a\}$.

In this reduction, the formula is copied as is in the theory
of the abduction problem. As a result, two formulae
$E|_{x=\true}$ and $E|_{x=\false}$ are translated into two
abduction instances differing only for the value of $x$ of
the theory $T$. We show that two such instances can merged
into a single instance with a moderate increase of size.

\begin{lemma}

For every $H$, $M$, $T$, the instance $\l H', M', T' \r$ has
exactly all explanations of $\l H, M, T|_{x=\true} \r$ with
$x^+$ added to each and all explanations of $\l H, M,
T|_{x=\false} \r$ with $x^-$ added to each, where variable
$x$ does not occur in $H$ and $M$, variables $x^+$, $x^-$,
and $q$ do not occur in $H$, $M$, and $T$, and:

\begin{eqnarray*}
H' &=& H \cup \{x^+,x^-\} \\
M' &=& M \cup \{q\} \\
T' &=& T \cup
\{x^+ \rightarrow q, x^- \rightarrow q,
x^+ \rightarrow x, x^- \rightarrow \neg x,
\neg x^+ \vee \neg x^- \}
\end{eqnarray*}

\end{lemma}

\proof Let $S$ be an explanation of $\l H',M',T' \r$. Since
$q \in M'$, and this variable only occurs in the clauses
$x^+ \rightarrow q$ and $x^- \rightarrow q$, then either $S
\cup T \models x^+$ or $S \cup T' \models x^-$. Since $x^+$
and $x^-$ does not occur positively in $T'$, this means that
either $x^+ \in S$ or $x^- \in S$, but not both, since
otherwise $S$ would not be consistent with $T'$.

The explanations containing $x^+$ are exactly the
explanations of $\l H, M, T|_{x=\true} \r$ with $x^+$ added
to each. Indeed, $x^+ \in S$ makes $S \cup T'$ equivalent to
$S \cup T|_{x=\true} \cup \{\neg x^-, q, x\}$, and
$T|_{x=\true}$, $H$, and $M$ do not contain $x^-$, $q$, and
$x$. Similarly, the explanations containing $x^-$ are
exactly the explanations of $\l H, M, T|_{x=\false} \r$ with
$x^-$ added to each. As a result, the set of explanations of
$\l H',M',T' \r$ is (apart from $x^-$ and $x^+$) the union
of the explanations of $\l H, M, T|_{x=\true} \r$ and of $\l
H, M, T|_{x=\false} \r$.~\qed

This lemma proves that two similar instances of abduction
can be combined into a single one having the union of their
explanations (apart from some variables added to each). As a
result, if one is able to translate two QBFs
$Q.F|_{x=\true}$ and $Q.F|_{x=\false}$, then one can combine
the resulting two instances into a single one that has
explanations if and only if $\exists x Q.F$. In order to
prove the hardness of the problem, one only needs to analyze
the increase of size due to merging.

\begin{theorem}[Alternative proof; originally proved by
Eiter and Gottlob \cite{eite-gott-95-a}]

The problem of explanation existence is \S{2}-hard.

\end{theorem}

\proof A QBF of the form $\forall Y. F$ is valid if and only
if the problem of abduction $\l \emptyset, \{a\}, \{\neg E
\vee a\}$ has explanations. This reduction has the property
that the matrix of the QBF is copied verbatim in the theory
of the abduction instance.

Let us now assume the existence of a reduction with the same
property from from QBFs having $Q$ as their sequence of
quantifier to abduction instances exists. If $F$ is a
formula made of variables of $Q$ plus $x$, one can apply the
previous lemma: $\l H', M', T' \r$ has explanations if and
only if either $\l H, M, T|_{x=\true} \r$ or $\l H, M,
T|_{x=\false} \r$ has, where $\l H, M, T|_{x=\true} \r$ and
$\l H, M, T|_{x=\false} \r$ are the results of translating
$Q.F|_{x=\true}$ and $Q.F|_{x=\false}$, respectively. As a
result, $\exists x Q.F$ is valid if and only if $\l H', M',
T' \r$ has explanations.

What remains to be proved is that iterating this process
does not generate instance of super-polynomial size. This is
in this case straightforward, as each merge only adds a
constant number of variables and binary clauses to the
instance.~\qed

In this proof, it may look like $\l H,M,T \r$ is an
abduction instance involved in the proof. However, it is
not. The instances used in the proof are $\l H',M',T' \r$,
$\l H,M,T|_{x=\true} \r$, and $\l H,M,T|_{x=\false} \r$. In
other words, $H$ and $M$ are meant to be used only with
$T|_{x=\true}$ and $T|_{x=\false}$ while $T$ is meant to be
used (with some formulae added to it) only with $H'$ and
$M'$.

More generally, the two instances corresponding to
$Q.F|_{x=\true}$ and $Q.F|_{x=\false}$ only contain the
formulae $F|_{x=\true}$ and $F|_{x=\false}$, respectively,
and not the formula $F$, which is instead contained in the
merged instance. In particular, $F$ contains the variable
$x$, which is not mentioned in the two QBFs $Q.F|_{x=\true}$
and $Q.F|_{x=\false}$; $x$ is the variable of the merged
instance that makes the merged instance become equivalent to
one of the two original ones when assuming the value $\true$
or $\false$.

 %

\section{Default Logic}

In this section, we show how a reduction from QBF to the
problem of skeptical entailment in default logic can be
raised by the addition of a universal quantifier. This
allows for an alternative proof of \P{2}-hardness of this
problem. Default logic has been introduced by Reiter
\cite{reit-80}; several variants have been proposed since
then \cite{luka-88,scha-92,anto-99,libe-07}. The problem of
checking whether a default theory skeptically entails a
formula is \P{2}-complete \cite{gott-92-b,stil-92}.


The \P{2}-hardness of a problem can be established by
showing a reduction from $\forall\exists$QBFs to the
problem. The starting point is a simpler reduction from
$\exists$QBF; this reduction is then raised by the addition
of universal quantifiers. The starting reduction is easy to
give: a propositional formula $E$ is satisfiable if and only
if $\l \{\frac{:a \wedge E}{a \wedge E}\},\emptyset \r$
skeptically entails $a$, where $a$ is a variable not
contained in $E$.

This is a reduction from $\exists$QBF to the problem of
skeptical entailment in default logic. The QBF is translated
in such a way its matrix only occurs once, as is, in the
resulting defaults. Two default theories obtained by
translating two QBFs having the same quantifiers and
differing only for the value of a variable in the matrix can
be merged as shown in the following lemma.

\begin{lemma}

For every set of defaults $D$ and variable $q$ not occurring
in $D$, the extensions of the following theory are exactly
the extensions of $\l D|_{x=\true}, \emptyset \r$ with $x$
and $p$ added to each and the extensions of $\l
D|_{x=\false}, \emptyset \r$ with $\neg x$ and $p$ added to
each.

\[
T=
\left\l
\left\{
\frac{:xp}{xp} ,~
\frac{:\neg xp}{\neg xp}
\right\}
\cup
\left\{
\left.
\frac{p \wedge \alpha:\beta}{\gamma}
\right|
\frac{\alpha:\beta}{\gamma} \in
D
\right\}
,\emptyset
\right\r
\]

\end{lemma}

\proof Since the first two defaults of $T$ are mutually
inconsistent, and they are the only ones that are applicable
to the background theory, each extension of this default
theory contains either $\{x,p\}$ or $\{\neg x,p\}$. The
extension of $T$ are therefore exactly the extensions of $\l
D, \{x,p\} \r$ and of $\l D, \{\neg x,p\} \r$. In turn,
these two theories have the same extensions of $\l
D|_{x=\true}, \emptyset \r$ and of $\l D|_{x=\false},
\emptyset \r$, apart from $x$ and $q$.~\qed

This lemma proves that two similar default theories can be
merged into a single one having the extensions of both.
Since skeptical entailment is considered, the latter theory
implies a formula $a$ if and only if both the two former
theories do. If the two theories result from translating
$Q.F|_{x=true}$ and $Q.F|_{x=\false}$, the merged theory
therefore entails $a$ if and only if $\exists xQ.F$ is
valid.

\begin{theorem}[Alternative proof; originally proved by
Gottlob \cite{gott-92-b} and Stillman \cite{stil-92}]

Skeptical entailment in default logic is \P{2}-hard.

\end{theorem}

\proof A formula $\exists Y.E$ is valid if and only if $\l
\{\frac{:a \wedge E}{a \wedge E}\}, \emptyset \r$
skeptically entails $a$. In this reduction, the matrix of
the QBF is copied verbatim in the justification of one
default, and does not otherwise affect the default theory.

Let us now assume that a similar translation from QBFs
having $Q$ as their sequence of quantifiers to skeptical
default entailment exists. We show a translation from QBFs
having $\exists xQ$ as their sequence of quantifiers.

Let $\exists x Q.F$ be such a formula. By assumption,
$Q.F|_{x=\true}$ and $Q.F|_{x=\false}$ can be translated
into two default theories where $F|_{x=\true}$ and
$F|_{x=\false}$ only occur as the justification of a single
default. As a result, the two theories corresponding to
$Q.F|_{x=\true}$ and $Q.F|_{x=\false}$ can be written as $\l
D|_{x=\true}, \emptyset \r$ and $\l D|_{x=\false}, \emptyset
\r$, respectively, for some set of defaults $D$. One can
then apply the previous lemma, which proves that these two
theories both skeptically entail $a$ if and only if $T$
skeptically entails $a$. In other words, $T \models a$ if
and only if $\exists xQ.F$ is valid. By construction, the
translation from $\exists xQ.F$ to $T$ has the same property
that the matrix $F$ is translated verbatim in a default.

In order to complete the proof, we calculate the increase of
size of the involved default theories when the addition of
quantifiers is iterated. This increase of size is that two
new defaults (of constant size) are introduced, and a
variable is added to the precondition of each default. As a
result, this step can be iterated with only a quadratic
increase of size, thus obtaining a reduction from
$\exists\forall$QBF to skeptical default entailment.~\qed

 %

\section{Planning}

The problem of establishing the existence of a plan in
STRIPS \cite{fike-nils-71} is \pspace-complete
\cite{back-nebe-92,byla-94}. We can use the method of
raising for proving the hardness of this problem. In this
case, we have to show that both $\exists x$ and $\forall x$
can be added to the front of a QBF while only producing a
constant increase of size in the corresponding planning
instance.

For the sake of simplicity, we consider an extension in
which the precondition of each action is a propositional
formula, rather than a list of positive and negative
literals. An action is therefore a pair $\l P,C \r$ where
$P$ is a formula and $C$ is a list of literals. This action
is executable if $P$ is valid in the current state; its
effect is to make all literals of $C$ valid.

Clearly, checking whether a formula containing no variable
is valid can be translated into a problem of plan existence.
Given such a formula $E$, just build the action $\l E,\{a\}
\r$, where $a$ is a (new) variable, and have $a$ being false
in the initial state and required to be true in the goal.
This instance has a plan (composed of a single occurrence of
the only action it contains) if and only if $E$ evaluates to
$\true$.

This is a translation from QBFs with no quantifiers to the
problem of planning. The matrix of the QBF is translated
verbatim in the precondition of one action; the goal is a
single variable. Let us now assume that such a translation
from QBFs having $Q$ as their sequence of quantifiers
exists, and prove the existence of a similar translation
from QBFs with a single more quantifier in the front.

In order to add an existential quantifier $\exists x$, we
add a new variable $p$ which is initially false, we add two
actions $\l \neg p, \{x, p\} \r$ and $\l \neg p, \{\neg x,
p\} \r$, and add $p$ as a precondition of all other actions.
This way, $p$ is required to be true before executing all
other actions, which is only possible if $x$ is made either
true or false; once $x$ has been given a value, $p$ is also
made true. This makes $x$ unmodifiable, because the first
two actions can no longer be executed, and no other action
makes $p$ false or changes the value of $x$. This way, a
plan exists if and only if the instance corresponding to
either $Q.F|_{x=\true}$ or $Q.F|_{x=\false}$ has a plan. The
increase of size is of two new actions, plus one more
literal in each action.

A universal quantifier is added as follows. Assume that $x$
is the variable we want to quantify, and that the goal $a$
is the postcondition of a single action. We add two new
variables $b$ and $p$, both false in the initial state, and
change the goal from $a$ to $b$. We also add $p$ as a
precondition to all other actions, and the following three
actions.

\begin{eqnarray*}
a_1 &=& \l \neg p, \{x, p\} \r \\
a_2 &=& \l a \wedge x, \{\neg x, \neg a\} \r \\
a_3 &=& \l a \wedge \neg x, \{b\} \r
\end{eqnarray*}

In the initial state, only the first action is executable.
It makes $x$ true and all other actions executable. Let $P$
be an irredundant plan of this instance. The last action of
$P$ is $a_3$, since is the only action that makes the goal
$b$ true. This action requires both $a$ to be true and $x$
to be false. This means that, at some point, $a_2$ has been
executed as well, since this is the only action that makes
$x$ false. As a result, we have that $P$ starts with $a_1$,
contains $a_2$, and ends with $a_3$. Let $P_1$ and $P_2$ be
the segments of $P$ between $a_1$ and $a_2$ and between
$a_2$ and $a_3$, respectively. We have that $P_1$ is a
sequence of actions that makes $a$ true while $x$ is true;
$P_2$ makes $a$ true while $\neg x$ is true. In other words,
$P_1$ and $P_2$ are plans for the case in which $x$ is
assumed true and false, respectively.

All considered reductions have the following two properties:
a.\  the matrix of the QBF only appears in the precondition
of a single action and does not affect the rest of the
planning instance; and b.\  the goal is a single variable.
Given a reduction with these two properties, one can create
a new reduction from QBFs with a single more quantifier in
the front.

We omit the formal proof of \pspace-hardness. We have
informally proved that both an existential and a universal
quantifier can be added to a QBF so that the corresponding
planning instance only gain moderately in size. In
particular, both changes add only two or three actions of
constant size, and only add a single precondition to all
other actions. This means that the instance that results
from adding $n$ quantifier has at most $3*n$ actions, and
each action is at most $m+n$ large, where $m$ is the size of
matrix.

 %

\section{Conclusions}

The method shown in this article allows for proving a result
of hardness for a class of the polynomial hierarchy without
directly building a reduction from the problem of validity
of a class of QBFs. The idea is that a proof of
\S{n}-hardness can be built by first showing that the
problem is \P{n-1}-hard, and then showing that the
particular instances used in this hardness proof are able to
``simulate'' an existential quantifier. In the same way, one
can prove \P{n}-hardness from a \S{n-1}-hardness proof. This
technique is shown here for explanation existence in
logic-based abduction (\S{2}-hard), for skeptical default
logic entailment (\P{2}-hard), and plan existence in an
extension of STRIPS (\pspace-hard); in a previous article,
it has been used to prove the \P{4}-hardness of a problem
related to redundancy in default logic \cite{libe-redu-3}.

This technique can be used in two ways: either for all
quantifiers or for the last one. For example, in the case of
planning we have shown that planning instance can always
simulate existential and universal quantifiers, therefore
raising a \S{0}-hardness result to \pspace-hardness. On the
other hand, the proof of \S{4}-hardness of the problem of
background theory redundancy in default logic
\cite{libe-redu-3} is based on first showing the problem
\P{3}-hard using a ``classical'' reduction from QBFs, and
then proving that existential quantifiers can be added,
therefore raising this result to \S{4}-hardness.

The main advantage of this technique is the simplification
of the proofs. While a proof of \S{4}-hardness would require
considering a QBF in the form $\exists X \forall Y \exists Z
\forall W.F$, raising an hardness result only requires to
prove that two similar instances of the problem can be
merged in the appropriate way. This means that considering
complicated QBFs may not be necessary.

 %

\let\sectionnewpage=\relax

\bibliographystyle{alpha}


\end{document}